\documentclass[sigconf]{acmart}
\usepackage{comment}
\usepackage{hyperref}
\usepackage{graphicx}
\usepackage{balance}
\usepackage[utf8]{inputenc} %
\renewcommand\footnotetextcopyrightpermission[1]{}

\title{Why Open Small AI Models Matter for Interactive Art}

\author{Mar Canet Sola}
\orcid{0000-0001-5986-3239}
\affiliation{%
  \institution{BFM, Tallinn University, Estonia}
  \city{}
  \country{}}
\affiliation{%
  \institution{Academy of Media Art Cologne(KHM), Germany}
  \city{}
  \country{}}
\email{mar.canet@gmail.com}

\author{Varvara Guljajeva }
\authornote{Both authors contributed equally}

\orcid{0000-0002-0261-3121}
\affiliation{%
  \institution{ VCUarts Qatar, Qatar}
  \country{}
}
\affiliation{%
  \institution{ Academy of Media Art Cologne, Germany}
  \country{}
}
\email{varvarag@gmail.com}

\setcopyright{none} %
\acmConference[]{Conference Name 'YY}{Month DD--DD, YYYY}{City, Country} %
\acmBooktitle{Conference Name 'YY: ACM Symposium on Something Something, Month DD--DD, YYYY, City, Country} %
\acmPrice{15.00} %
\acmDOI{10.1145/xxxxxxx.xxxxxxx} %
\acmISBN{978-x-xxxx-xxxx-x/YY/MM} %

\begin{abstract}
This position paper argues for the importance of open small AI models in creative independence for interactive art practices. Deployable locally, these models offer artists vital control over infrastructure and code, unlike dominant large, closed-source corporate systems. Such centralized platforms function as opaque black boxes, imposing severe limitations on interactive artworks, including restrictive content filters, preservation issues, and technical challenges such as increased latency and limiting interfaces. In contrast, small AI models empower creators with more autonomy, control, and sustainability for these artistic processes. They enable the ability to use a model as long as they want, create their own custom model, either by making code changes to integrate new interfaces, or via new datasets by re-training or fine-tuning the model. This fosters technological self-determination, offering greater ownership and reducing reliance on corporate AI ill-suited for interactive art's demands. Critically, this approach empowers the artist and supports long-term preservation and exhibition of artworks with AI components. This paper explores the practical applications and implications of using small AI models in interactive art, contrasting them with close source alternatives.
\end{abstract}

\keywords{AI models, Small models, Open Source, Interactive art, AI art, Real-time systems, Creative autonomy, Art preservation}

\begin{document}
\maketitle

\section{Introduction}

In recent years, artificial intelligence has significantly expanded the creative possibilities of creative fields such as digital art, film, music, design, and, notably, interactive art \cite{epstein2023art, manovich2022ai, kelomees2022meaning}. Artists and designers are increasingly incorporating AI into their workflows, but often relying on proprietary platforms. AI methods might allow new creative possibilities, but they are not magic and have many limitations that might consume the same amount of time as before for creatives. For example, Sora's video generation tool in 2023 was received with mixed reactions from filmmakers, with many reporting that only one of the 300 generated clips might be considered usable \cite{Meyer2024} in specific tasks. These proprietary AI platforms offer easy-to-use systems to attract more users, but hide severe limitations. As Abuzuraiq and Pasquier argue, large-scale closed-source models limit users to simple prompting rather than genuine interaction, reducing their ability to shape AI tools to their specific needs \cite{abuzuraiq2024towards}. 

In addition, big AI brands used artists for advertising and 'art-washing' campaigns, leveraging their work to promote their proprietary AI models without providing real support or fair compensation. A striking example is OpenAI's Sora, which granted selective free access to a handful of artists solely to market the tool while heavily controlling the published outcomes. This lack of transparency and genuine collaboration led to significant backlash from the artistic community, culminating in the petition "Dear Corporate AI Overlords,",\footnote{\url{https://openletter.earth/dear-corporate-ai-overlords-90668a95}} , where artists demanded their rights and called for the decolonization of AI technology. This idea of decolonization is one of the inspirations for using small open AI models for the artist.

By small open AI models, we mean locally runnable models whose weights are openly redistributed under permissive licenses. This concept resonates with calls for more artist-centric approaches, such as the notion of 'human-scale models' proposed in the XAIxArts Manifesto \cite{bryankinns2025xaiarts}, which emphasizes the importance of the manageability and accessibility for creative practitioners of these technologies.

The AI industry's reliance on high-performance GPUs exemplifies this growing divide of power and raises concerning issues. These types of GPU are prohibitively expensive for individual artists and even some academic institutions. For instance, NVIDIA's H200 GPUs, released in 2024, costs up to \$40,000 per unit, reinforcing the concentration of AI power among a few well-funded corporations.\footnote{\url{https://www.forbes.com/sites/bethkindig/2024/06/20/ai-power-consumption-rapidly-becoming-mission-critical/}} Additionally, as AI models grow in size require multiple and larger GPUs to run the training and inference of these AI models, this approach necessitates increasingly energy-intensive hardware, exacerbating environmental concerns. The NVIDIA A100 GPU, released in 2020, consumed up to 400W, while its successor, the H100, released in 2022, increased power consumption to 700W—a 75\% increase in just two years. Although the H200 maintains the same power consumption as the H100, it does so while delivering significantly higher performance, effectively reducing energy use per computation. However, the growing dependence on large-scale artificial intelligence still contributes to the escalating global electricity demand, intensifying concerns about sustainability.

Beyond environmental and financial constraints, the dominance of proprietary AI platforms raises deeper structural issues. These cloud-based services operate as black boxes, offering little transparency about their inner workings (like energy consumption and water), data sources, or biases. Artists and designers lack deep control over these tools and rely on corporations that can determine the inference price, restrict features, ban topics, and discontinue models based on commercial strategies. This dynamic reflects technological centralization, where creative practitioners must conform to pre-built AI models rather than shaping the technology to suit their specific artistic needs, hindering the development of novelty in artistic practices.

Unlike many large models, some smaller open-source models can be run on personal devices (often on consumer-grade GPUs and sometimes on CPUs). These models, including generative and classification AI, allow interactive artists to fine-tune tools with curated datasets, integrate them with various sensors and physical computing setups to develop new forms of input, create novel interfaces, and ensure their artworks remain functional without relying on external corporate services.

The shift toward open-source, small-scale AI aligns with broader movements advocating for technological self-determination. Open foundational models such as Stable Diffusion (image generation), YOLO (object detection with segmentation variants), BERT and CLIP (text and multimodal encoders), Whisper (speech-to-text) and large language models such as BLOOM, Llama, Gemma, and Mistral (general-purpose language modeling) demonstrate how decentralization can empower individuals to develop, modify, and sustain impactful AI tools \cite{kapoor2024societal}. Open small AI models provide a pathway toward more ethical, sustainable, and democratized AI practices, particularly within interactive art, a medium where software is executed live during exhibitions rather than presented as a static, pre-recorded form like video or sound.

\section{Open Small AI Models for Interactive Artists and Designers}

The concept of a \textit{small model} refers only to computational size. Ideally, the size of a \textit{small model} should be compact enough to run inference on a single machine, whether on an artist's local device. However, for these creative uses, it also needs to be open licensed with redistributable weights because this combination enables modification, preservation, and a gain in artistic autonomy.

Notable examples include variants of Stable Diffusion for image generation optimized for realtime like StreamDiffusion \cite{hacohen2024ltx} \footnote{https://github.com/cumulo-autumn/StreamDiffusion}, video models like LTX-Video \footnote{https://huggingface.co/Lightricks/LTX-Video} or MotionStream \cite{shin2025motionstreamrealtimevideogeneration} and small LLMs, which can run on gaming computers typically equipped with NVIDIA GPUs running Linux or Windows, or on devices Apple computers, widely used within the creative community. Running small models locally can achieve the real-time and low-latency response essential for interactive works. %

Artists have already begun integrating small models into their creative workflows, demonstrating the power of local and adaptable AI systems. Evidence from practice includes Varvara \& Mar's interactive installations, such as \textit{Dream Painter} (2021) \cite{guljajeva2022dream} and \textit{Vision of Destruction} (2023) \cite{canet2024visions}, where custom AI pipelines run on local machines, ensuring the real-time responsiveness and complete control necessary for direct audience interaction \cite{guljajeva2024artist, canet2024visions}. 

A different non-interactive example is Superradiance (2024) \footnote{\url{https://superradiance.net/\#about}}
, a collaboration artwork between Memo Akten and Katie Peyton Hofstadter that was developed using small open-source AI models within TouchDesigner, ComfyUI and coding, creating a custom pipeline for each chapter \footnote{A detailed explanation of the making-of \textit{Superradiance}: \url{https://www.youtube.com/watch?v=B_igdUDzcs4&t=511s}}. This allows for a highly controllable approach to AI-driven artwork, resulting in an immersive multichannel video installation that explore embodiment, technology exploration, and planetary consciousness  . 

\begin{figure*}[t]
  \centering
  \includegraphics[width=0.9\linewidth]{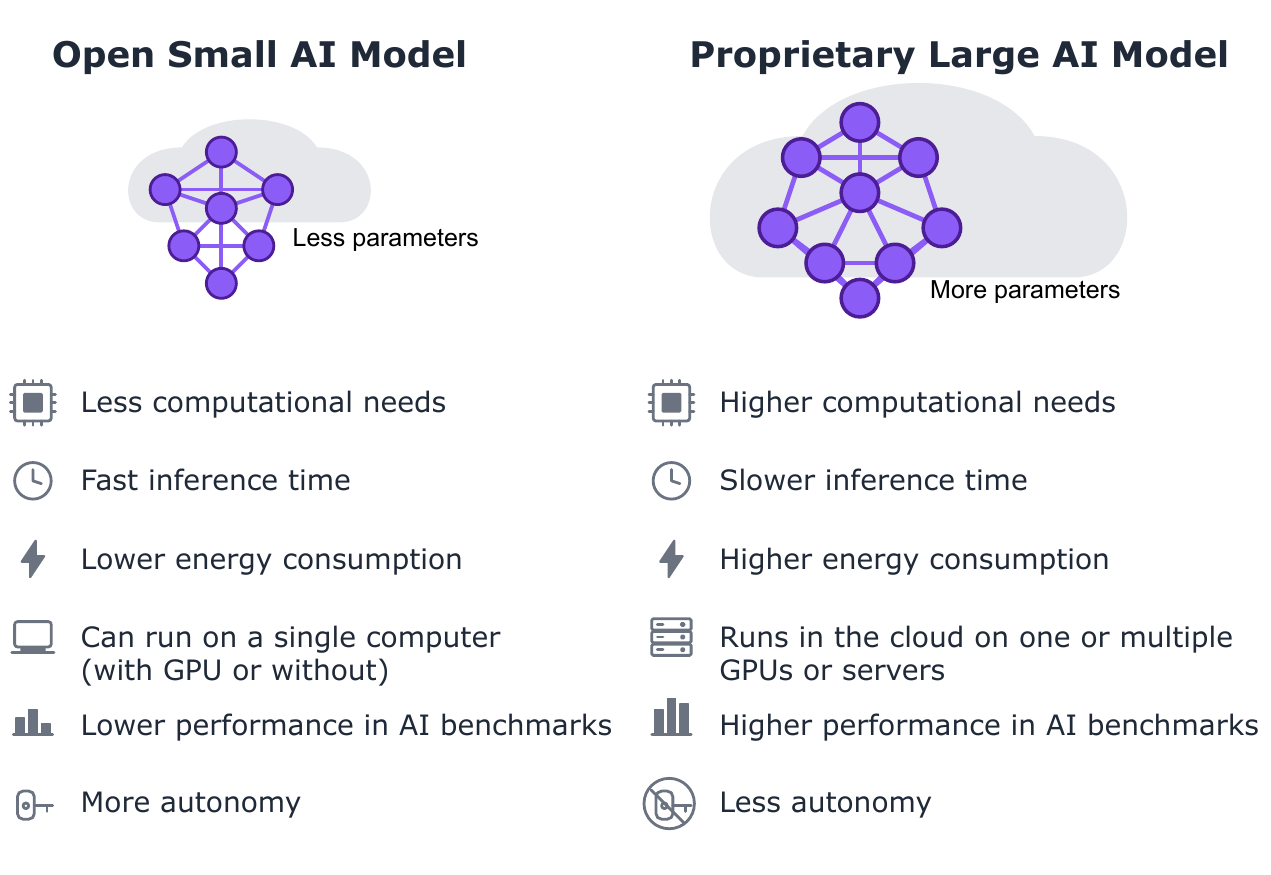}
  \caption{Comparison of small and large AI models across key characteristics. }
  \label{fig:comparison}
\end{figure*}

A broader comparison between small and closed source large AI models illustrates the trade-offs involved in choosing between the two. As shown in Figure \ref{fig:comparison}, large-scale proprietary models can excel in better performance due to their extensive training in larger datasets. However, this advantage comes at a cost, both financially, environmentally, and with less autonomy. Large models require substantial computational infrastructure, which increases operational expenses and energy consumption. Cloud-based AI services, while accessible to users via simple interfaces, often come with hidden costs such as data restrictions, subscription fees, and the risk of discontinuation of services if a company decides to phase out a particular tool.

\subsection{Core Characteristics of Open Small AI Models}

\begin{itemize}
    \item \textbf{Open Source:} Most open AI models are small. Open models are the most suitable for interactive art because they are available for inference and modification, ensuring artists can maintain and adapt their interactive systems long-term.
    \item \textbf{Selectable Models and Checkpoints:} Open-source models offer users extensive control, allowing them to use older or newer models version and select among different checkpoints (training files used for inference). This is vital for interactive art, allowing artists to design the experience as they want with a specific model version used in a piece, and ensuring preservation. The open models foster diversity, allowing users to choose from various model flavors, each with slight adjustments tailored to specific creative needs. For example, repositories like Civit.ai \footnote{\url{http://civit.ai}} that contain a large number of community-driven checkpoints for image models provide nuanced options, allowing users to precisely choose a specific style that can be used in their creations. 
    
    \item \textbf{Single-Machine Operability / Local Execution:} Small models are optimized for individual hardware, reducing dependency on cloud infrastructure. This enables offline operation, which is essential for installations in diverse locations (good and stable internet can sometimes be hard to get in some exhibition venues) and provides a lower latency required for seamless real-time interaction. Creating a simpler setup that is free from external network complications and expensive API charges.
    \item \textbf{Enhanced Control and Ownership:} By running models locally, interactive artists have significantly greater control over hardware, software, integration with sensors/physical elements, and creative outcomes. 
    \item \textbf{Sustainability:} Small models with fewer parameters, capable of running on local machines, consume significantly less energy than large-scale models typically hosted in the cloud \cite{luccioni2024power}. This makes them a more sustainable choice for creators of environmental concerns. Furthermore, these models can be easily monitored on local machines, providing detailed insight into resources being consumed, such as electricity usage. Enabling artists to make more informed and empowered decisions in their artwork design.
\end{itemize}

Adopting open small models is not just a technical shift, but can be seen as a cultural and ethical movement that redefines how AI is integrated into creative practices. By emphasizing decentralization, sustainability, and user autonomy, small models challenge the dominant AI paradigm and propose a future where technology serves artists rather than the other way around. The choice between centralised and decentralised models will shape the next generation of creative tools as AI evolves. By embracing small models, artists and designers can actively shape this future—one where innovation is driven by independence rather than dictated by corporate AI providers.

\section{Advantages of Open Small AI Models for Interactive Art and Shortcomings of Proprietary Large AI Models}

For interactive art, one of the core advantages of open small AI models running locally is when real-time responsiveness is achieved. Unlike cloud-based models, which introduce network latency, locally operated small models enable instant interaction—an often critical requirement for installations and performances that engage directly and immediately with audiences or environmental input. Even minor delays in cloud-based systems can disrupt fluid responsiveness, undermining the core of the interactive experience. Small models are therefore highly suitable for interactive environments where immediate feedback is crucial.

The study by Canet Sola and Guljajeva (2024) illustrates this through \textit{Visions of Destruction}, an interactive artwork that leverages generative AI models operating locally to provide seamless audience interaction \cite{canet2024visions}. The critical importance of low-latency processing is further highlighted by Kyle McDonald's installation \textit{Transformirror} (2023), showcased at the KIKK Festival (2024) in Belgium. McDonald's research found that despite using a high-performance RTX 4090 GPU in Romania through RunPod, cloud-based inference still resulted in significant latency (70-74 ms per frame, with an additional 100 ms delay for diffusion image generation) \cite{mcdonald2024realtime}. This resulted in a significant drop in performance, reducing the expected 10 frames per second to 5.7 FPS, demonstrating that cloud-reliant AI often struggles to meet the fundamental demands of real-time generative interaction. These findings underscore the potential of small models that run locally for interactive and performance-based works.

When AI models are an integral part of the software pipeline, artists using cloud-based solutions risk catastrophic failure if providers discontinue services, could face changes due to API versions, and restrict certain functionalities that are only possible when you run locally, such as integrating custom input interfaces (e.g. sensors or other peripherals). An interactive installation intended for long-term gallery display can face significant risks relying on such unstable foundations. Locally hosted, open-source models ensure the longevity and maintainability of AI-integrated artworks, allowing creators to archive their models and workflows, guaranteeing future operation independent of shifting commercial policies. This self-sufficiency is particularly vital in media and interactive art, where the preservation of original tools, behaviors, and outputs is integral to the integrity of the art over time \cite{lozano2022best}. Furthermore, the ability to understand and maintain these models contributes to what Tecks et al. (2024) describe as 'explainability paths for sustained artistic practice,' allowing artists to engage with their AI-driven works over extended periods \cite{tecks2024explainability}.

One case in which it is appropriate to use cloud-based AI servers, particularly when they can be self-hosted using open small models, is in interactive net-art. An example is \textit{What do you want me to say?} by Lauren Lee McCarthy \footnote{\url{https://get-lauren.net/What-do-you-want-me-to-say}}
. In this work, the artist creates a digital clone of her own voice, effectively becoming a puppet that repeats what participants tell her to say. After asking “What do you want me to say?”, the cloned voice speaks to the participant’s words back to them. The work was originally created in 2021. If the piece were not based on an open small AI model, it would be difficult to show again years after its creation, as seen in its exhibition in 2025 at the Festival de Arte Digital da Trafaria, Periphera (Lisbon). 

Another net-art artwork, \textit{Keep Smiling} \footnote{\url{https://var-mar.info/keep-smiling/}}
 by Varvara \& Mar, simulates a job interview with an AI agent that instructs the candidate to keep smiling while continuously evaluating the quality of their smile as part of an endless interview process \cite{guljajeva2022keepsmiling,guljajeva2022keep}. The work is both humorous and critical, offering a playful yet pointed commentary on current practices that involve the use of AI in evaluating job candidates. Moreover, this piece exemplifies how a small AI model can run directly in the browser without relying on external servers, as the model is lightweight enough to operate on participants’ regular devices while they interact with the artwork through their webcams.

Another notable example is the long-term collaboration of the Berlin-based painter Roman Lipski’s\footnote{\url{https://www.romanlipski.com/}} with his \textit{AI Muse} developed using the pix2pix Generative Adversarial Network (GAN) model. The Muse tool was developed in 2016 in collaboration with Florian Dohmann, the programmer. Muse features an innovative interface: a grid of knobs that enables the artist to mix different models, each trained with previous paintings by Lipski. This system generates a high number of permutations and possibility spaces allowing for emergent creative explorations by the artist, which he had used for his painting studio creations, but also in performances and workshops. The process involved integrating AI as an active co-creator and source of inspiration, treating the entire generative pipeline (software, code, and the original Lipski painting dataset) as part of the artistic legacy to be preserved and reactivated in future exhibitions, whether as performances or as a means of showing the artist’s process \footnote{\url{https://ai-berlin.com/blog/article/interview-with-roman-lipski-ai-artist}}. This highlights the importance of not only archiving the final product, allowing it to be used and explored by the artist over many years, and ensuring preservation of the artistic processes.

As seen in the Lipski example, a major benefit of small models is their flexibility and customization. Unlike proprietary cloud-based AI, which imposes fixed parameters and limited modification options, small models enable artists to fine-tune AI behavior to align with specific artistic styles. This can be achieved through custom checkpoints, dataset curation, manipulation of model parameters, or additional training rounds, giving creators enhanced control over how their AI generates outputs or responds to inputs within an interactive loop. This allows for the creation of unique interactive personalities or responses tailored to the specific artistic concept \cite{vigliensoni2022small}. Such deep engagement moves beyond simple prompting, allowing artists to truly 'seize the means of production,' as Abuzuraiq and Pasquier (2024) articulate, by actively crafting, adapting, and navigating generative models to suit their expressive needs \cite{abuzuraiq2024seizing, abuzuraiq2024towards}. A compelling example is Helena Sarin, who personally trained her AI models using a curated dataset of her own artworks.\,\footnote{\url{https://www.nvidia.com/en-us/research/ai-art-gallery/artists/helena-sarin/}} By feeding the AI her drawings, sketches, and photographs, she transforms the model's output into a highly personalised aesthetic, blending algorithmic generation with human artistic intent. Her work showcases how small models empower artists to develop unique machine-learning tools rather than relying on generalised, one-size-fits-all AI systems. Similarly, \textit{xhairymutantx} (2024)\,\footnote{\url{https://xhairymutantx.whitney.org/}} by Holly Herndon and Mat Dryhurst explores AI-generated identity through a fine-tuned text-to-image model that allows audience interaction to shape Herndon's evolving digital persona. \textit{xhairymutantx} is an example of an interactive net-art artwork using AI, where the outputs are co-created by the audience, the artist, and the AI system, with the aim of injecting these generated images into future training sets, thereby transforming the representation of Holly within them. Furthermore, these cases illustrate how custom small models facilitate deep artistic exploration and participatory experiences in ways that corporate AI tools often cannot for interactive cases.

Another key advantage of small models is their resilience against commercial restrictions. Cloud-based AI services are inherently vulnerable to policy changes, model discontinuations, and paywalled access. OpenAI's decision to retire DALL·E 2 despite user demand demonstrates how companies control access to AI tools, often prioritizing profit over artistic utility  \footnote{\url{https://www.pcmag.com/news/openai-shuts-down-dall-e-2-image-generator}} . For an interactive artist mid-project or maintaining an installation, such a decision can be devastating. In contrast, locally deployed models offer stability and independence, ensuring that artists are not forced to abandon or drastically rework functioning interactive systems due to corporate decisions beyond their control.

Privacy and intellectual property protection are also significant concerns with cloud-based AI. Many online AI services require users to upload creative data, which raises concerns about data ownership, algorithmic bias, and unauthorised use. By contrast, running AI models locally means all data (potentially sensitive audience interaction data) stays on the artist's machine, ensuring full control over training datasets, generated outputs, and creative processes.

The autonomy granted by small models extends beyond individual projects to broader creative freedoms. Cloud-based AI services enforce content moderation policies that restrict certain themes, such as nudity, depictions of public figures, or politically sensitive material. While these regulations may serve corporate interests, they can limit artistic expression, particularly in experimental, critical, or activist art, including critical or experimental interactive works that might explore sensitive themes through audience participation. With small models, artists retain greater agency over their AI-generated works, free from corporate censorship.

Crucially, this freedom extends to the ability to 'pervert technological correctness,' as Rafael Lozano-Hemmer termed it \cite{lozano1996perverting} – the artistic impulse to misuse technology, explore its limitations, and push it beyond its intended functions, rather than adhering strictly to a corporate-defined 'correct' usage. Large, closed systems, often designed to prevent unexpected outputs and enforce specific norms, inherently restrict this type of critical and experimental engagement. Open small models, by providing full control and modifiability, empower artists to engage in such 'perversion,' which can be a vital source of artistic innovation and critique.

Beyond artistic control, small models also align with sustainability efforts. Unlike large, cloud-hosted models, which demand high-powered GPUs and extensive cooling infrastructure, small models run on comparatively energy-efficient local hardware. AI data centers contribute significantly to global energy consumption. In contrast, small models can operate on a fraction of this energy, reducing the ecological footprint and operational cost, making long-term interactive installations more feasible \cite{luccioni2024power}.

While cloud-based AI offers convenience through pre-trained, large-scale models with simple interfaces, this ease of use comes at the cost of user control, flexibility, real-time capability, and long-term accessibility/stability. Subscription fees, access restrictions, data regulations, and unpredictable service lifespans further limit the independence of artists, making cloud-based AI a challenging and often unsustainable choice for serious engagement in interactive art. In contrast, small AI models empower users to work on their own terms, ensuring customisation (allow code base modifications to the model, training or fine-tuning), privacy (avoid privacy issues with personal data of participants), responsiveness (reduction latency), stability(allowing preservation of artworks), and sustainability. As AI becomes more deeply embedded in creative workflows, choosing a decentralized and local approach to run AI models, artist-driven technologies will be key to maintaining artistic freedom and ensuring the viability of interactive art in the digital age.

\section{Ownership, Accessibility, and Control: Open-Source Licensing in AI}

Open-source licenses play a fundamental role in shaping the accessibility, usability, and longevity of AI models for creative professionals. For AI models to best support interactive artists, they should ideally be released under permissive open-source licenses, such as MIT, Apache, or CC0, which guarantee freedom in modification, distribution, and commercial use. These licenses allow artists to integrate AI tools into their workflows without restrictive legal barriers, ensuring that AI models remain adaptable, shareable, maintainable, and archivable for the long lifespan often required by interactive artworks.

However, many models labeled as "open-source" do not fully meet this standard. Some AI developers open-source only the codebase but keep training checkpoints proprietary, significantly limiting practical usability. Without access to training checkpoints, artists cannot fine-tune or modify models meaningfully, restricting the customisation needed to tailor AI behaviour for specific interactive scenarios or the ability to guarantee the long-term function of an artwork. A genuinely open AI model should include both the source code and the training checkpoint; ideally, the training dataset should also be available. This transparency fosters innovation, allowing artists to build upon existing models, modify outputs, and ensure their interactive systems can be preserved and restaged.

Despite the clear advantages of openness, some AI models adopt restrictive licensing policies that hinder creative professionals. For instance, the FLUX.1 [dev] Model, developed by Black Forest Labs Inc., is licensed under the FLUX.1 [dev] Non-Commercial License, explicitly prohibiting all commercial use.\,\footnote{Licensed under the FLUX.1 [dev] Non-Commercial License by Black Forest Labs Inc. For more information, see: \url{https://github.com/black-forest-labs/flux/blob/main/model_licenses/LICENSE-FLUX1-dev}} This restriction makes the model inaccessible to artists who wish to exhibit or potentially sell interactive works incorporating it, reducing its utility for independent creators, freelancers, and small studios. Such licensing structures conflict with the principles of open-source development by restricting how artists can legally use and distribute AI-generated work and hinder the sustainability of artistic practices that rely on these tools.

A broader issue facing artists working with AI is the forced transition to newer models due to corporate decisions. Many commercial AI providers frequently discontinue older versions of models for financial and strategic reasons, rendering previous tools unusable. This practice disrupts established creative workflows and significantly threatens the preservation and continued operation of digital and interactive art. In interactive and media arts, where AI models function as an integral part of the software pipeline and often interact with custom hardware/sensors, such disruptions can lead to the loss of critical functionalities, forcing artists to re-engineer their works from scratch, often an impossible task if original behaviours cannot be replicated. When proprietary models are shut down or updated without backward compatibility, interactive artworks may become permanently non-functional, making long-term project sustainability highly uncertain.

In the context of media art conservation, AI model longevity, and the specific challenges of preserving interactive systems, Rafael Lozano-Hemmer's work \textit{Best Practices for Conservation of Media Art from an Artist's Perspective} offers valuable insights \cite{lozano2022best}. Lozano-Hemmer argues from extensive artistic practice that artists must maintain control over their digital tools to ensure the longevity and integrity of their works, advice that is paramount for interactive art. He advocates for open-source frameworks to preserve media art, aligning with the broader philosophy of transparency, communal improvement, and digital self-sufficiency. His recommendations support the argument that open-source is often the best option for preserving complex, technology-dependent interactive artworks, as they allow artists to continuously update, modify, adapt, and archive their tools without fear of them becoming obsolete due to corporate decisions. This reinforces the notion that open AI models share the same preservation and autonomy advantages as other open-source software components within artistic workflows.

Beyond preservation, the benefits of open-source AI models extend to artistic freedom and long-term accessibility. Open models enable community-driven innovation, where artists, designers and developers can collaborate in building communities to refine AI tools, share knowledge, and expand creative possibilities \cite{broad2021active}. This decentralized approach reduces the reliance on proprietary infrastructures, fostering an environment where AI can be used and adapted for specific interactive needs without financial or contractual constraints. Open models also encourage more sustainable AI practices, allowing reuse, optimization, and redistribution rather than forcing users to adopt resource-intensive new models dictated by corporate release cycles, ensuring interactive art can thrive outside the dictates of corporate release cycles.

In conclusion, choosing open-source AI models is not just a technical decision but an artistic requirement for interactive artists that need to ensure creative control, long-term viability, and the archival integrity of their work. Many artistic strategies used in interactive art using GenAI will not be possible if an artist does not have access to source code, training checkpoints, and ideally, training datasets. This choice ensures that artists retain creative autonomy, preserve their work over time, and remain free from the restrictions of corporate-controlled AI infrastructures. As the integration of AI into artistic workflows continues to evolve, advocating for fully open and permissive AI models will be key to ensuring that AI remains a tool for artistic empowerment rather than a means of technological dependence.

\section{Discussion}

Although open small AI models offer significant advantages for interactive art, their adoption also presents challenges that must be considered. These challenges revolve primarily around hardware limitations, the costs of equipment, and technical expertise. However, as open-source communities grow and AI tools become more user-friendly, many of these obstacles are gradually being addressed, making small models increasingly accessible to interactive practitioners.

One of the main barriers to entry is hardware limitations. Although small models are designed to be lightweight compared to their large-scale cloud-based counterparts, they still often require high-performance GPUs to run efficiently. This poses a challenge for interactive artists lacking access to expensive hardware, particularly in developing regions with higher technology costs, especially when needing reliable, dedicated machines for installations. Limited access to advanced computing power can restrict creative possibilities and exclude certain artists from AI-driven artistic exploration. However, recent advancements in optimization techniques (e.g. quantization like GGML/GGUF, model pruning, or knowledge distillation) have made running small models on less powerful machines possible. These improvements help lower hardware requirements for AI interactive art pieces, making it more inclusive and reducing the technological gap between artists with different levels of access to hardware resources.

Unlike proprietary AI tools that often provide ready-to-use solutions, open small AI models typically require manual installation, configuration, integration into larger interactive systems, and ongoing maintenance. This complexity can discourage artists without programming experience from working with open-source AI. However, recent approaches such as vibe coding, supported by large language models, have begun to reduce some of this complexity. However, community-driven initiatives and user-friendly open-source tools are actively lowering these barriers. Platforms like ComfyUI \footnote{ComfyUI: A highly modular and powerful diffusion model interface that offers a GUI, API, and backend with a graph/nodes structure for flexible model customization: \url{https://github.com/comfyanonymous/ComfyUI}} provide intuitive graph-based user interfaces, allowing artists to experiment with AI models without extensive coding knowledge. Additionally, the availability of step-by-step tutorials on platforms like YouTube and different communities fosters a collaborative learning environment, allowing artists to acquire basic technical skills and integrate AI into their creative workflows. The availability of tutorials on integrating AI with platforms like Processing, OpenFrameworks, TouchDesigner, or microcontrollers fosters a collaborative learning environment.

Other essential considerations are model maintenance and updates, especially for long-running interactive installations. Unlike cloud-based services that automate updates and improvements, artists using small models must manually update and maintain their tools. This can be a burden, particularly for independent artists who lack the time or technical expertise to adapt to new software changes continuously. Furthermore, frequent updates in open-source AI models can disrupt stable interactive systems, requiring users to reconfigure their pipelines. However, active open-source communities help mitigate this issue by providing long-term support, troubleshooting guides, and collaborative development. Platforms like ComfyUI foster environments where users collectively maintain tools, ensuring that even artists who are not deeply technical can benefit from ongoing improvements, offering some stability, although careful version management by the artist remains crucial for long-term projects. The capacity of artists to deeply understand and manage their chosen models, a benefit often afforded by smaller open systems, can also support the "sustained artistic practice" advocated by Tecks et al. (2024), by fostering a more profound and enduring relationship with the technological medium \cite{tecks2024explainability}.

To address some of these challenges, hybrid cloud solutions that still provide some degree of user control offer a middle ground. Services like Google Colab, Dockerized environments, and locally controllable cloud GPU instances (like RunPod, Vast.ai) enable artists to access powerful computing resources on demand without requiring local high-end hardware, while potentially still allowing local control over the model/code execution for lower latency than pure SaaS APIs. These cloud solutions operate pay-per-use, allowing independent creators to run more complex models affordably while retaining control over their tools and workflows. Balancing cloud accessibility with open-source flexibility, these solutions help bridge the gap between creative autonomy and computing power, though they may reintroduce some dependency and latency compared to fully local setups.

The shift toward small, locally-run AI models can be seen not just as a technical change—it arguably represents a transformation towards reclaiming creative independence from dominant corporate AI platforms. This aligns with the idea of artists 'seizing the means of production' by gaining deeper control over the crafting and adaptation of their AI tools (Abuzuraiq and Pasquier, 2024) \cite{abuzuraiq2024seizing}. The ability to modify, control, and customise AI tools aligns with the values of many interactive artists and designers who prioritise autonomy, experimentation, and direct audience engagement. As the adoption of small models increases, we may witness the emergence of niche, community-driven AI tools that cater to specific artistic styles, interactive behaviours, and experimental approaches, creating a more diverse and decentralised AI landscape suited to varied interactive contexts.

This decentralisation helps prevent monopolisation in creative AI tools. Large corporations currently dominate AI development, shaping how artists interact with technology by imposing restrictive terms of use and access limitations often incompatible with the needs of interactive art (latency, stability, integration). The rise of small models challenges this top-down approach, ensuring that AI remains an accessible, adaptable, and reliable medium for interactive creation. Small models foster a pluralistic creative ecosystem where unique perspectives and regional artistic traditions can be represented rather than filtered through corporate-driven AI aesthetics or constrained by rigid API limitations by allowing artists to fine-tune and personalise their AI tools.

Open-source AI's collaborative and decentralised nature also encourages cross-cultural exchange in digital art. As more artists engage with these technologies—and some contribute to their development and refinement—open-source communities around tools like Stable Diffusion Web UI and ComfyUI facilitate knowledge sharing across geographical and cultural boundaries. These communities empower artists from diverse backgrounds to actively shape the evolution of AI, ensuring that it reflects a broad range of artistic voices rather than reinforcing dominant Western-centric aesthetics. Furthermore, the instability of proprietary AI services presents a major challenge for artists relying on corporate models. Many companies frequently discontinue models or restrict access to older versions, forcing users to migrate to newer, often more expensive solutions. This lack of backward compatibility can disrupt long-term projects, particularly in fields like interactive and media art, where the AI model itself is embedded into the software pipeline. When proprietary models become unavailable or are altered, artists are often forced to redesign their entire creative process, which may compromise the original intent of their work. In contrast, small open-source models offer the long-term stability essential for preserving and exhibiting interactive art.

An interesting consequence of the rapid evolution of AI technology is that older AI models considered outdated by industry may later be rediscovered and repurposed for artists looking for unique aesthetics. Repositories like Huggingface models \footnote{https://huggingface.co/models} are incredible reserves of this huge number of AI models ready to be explored by artists. Ultimately, small AI models encourage experimentation in ways that proprietary tools often do not. Because artists can freely modify, tweak, integrate, and customise these models, they foster a culture of innovation where creators are not merely consumers of AI tools but active participants in developing unique interactive systems. This creative agency ensures that AI remains an open-ended medium for interaction, rather than a rigid, standardised tool dictated by corporate interests.

Moreover, decentralising AI development amplifies artistic voices that might otherwise be marginalised. Corporate AI models often enforce content restrictions based on economic and political considerations, limiting artistic expression in activist art, socially critical works, or experimental interactive digital aesthetics that might challenge norms. Small models remove these constraints, ensuring that AI can serve as a platform for radical creativity, self-expression, and cultural critique.

By allowing artists to develop their own AI-driven tools, particularly for interactive art, open small models ensure that AI does not become a static, commercialised product but remains a dynamic, evolving medium shaped by the communities that use it. This shift represents a move away from passive AI consumption toward active AI creation, where artists are no longer just users of AI but co-developers of the tools that shape their art.

Finally, as educators in art institutions teaching courses on interactive art and creative generative AI, you face an important choice: you can either purchase subscriptions to proprietary tools or invest in infrastructure by acquiring computers capable of running open-source models.

The first option, subscriptions, requires ongoing payments and in many cases these services use a credit-based system. This means that if a student wants to pursue a project, they might quickly run out of credits and be forced to pay more. The second option, running open source models locally, empowers students and increases their knowledge, as they have full control over AI, computer, and all aspects of the technology stack (gaining control over the creative infrastructure). Although open source models might not always match the quality of proprietary ones, they offer more flexibility and can be adapted to better suit all creative needs.

It is also important to note that for those outside the field, there is a significant difference between buying a server, running AI in the cloud, and running it locally. For interactive art, running locally often makes more sense, as input devices, such as cameras or sensors, are typically connected to the same computer. Simple factors like internet connectivity can become complicated in exhibition spaces, and if your piece relies on an external API, a stable internet connection, and a computer, the chances of one of those components failing increase. Unless your interactive art is a net-art that uses AI, then you will need an AI server.

\section{Conclusion}

Open Small AI models present a compelling and often necessary alternative to centralized commercial AI services, offering interactive artists greater creative autonomy, sovereignty, flexibility, customization, real-time performance, and long-term stability. Open small models prioritize artistic exploration and independent innovation, unlike proprietary AI models, which operate within profit-driven frameworks often incompatible with the demands of artistic practices like interactive art. By providing enhanced control over training, fine-tuning, integration, and deployment, artists can shape AI as a tool that aligns with their personal vision and the specific needs of their interactive systems, rather than being constrained by corporate interfaces, policies, and the inherent instability of closed platforms.

While large AI models developed by major tech companies will continue to be popular due to their convenience and power for non-real-time tasks, small models offer a different and often essential value for interactive contexts—one rooted in self-sufficiency, transparency, customization, low latency, and longevity. Their typically open-source nature fosters collaboration and knowledge sharing, creating a more decentralized and diverse AI ecosystem that empowers creators of interactive experiences rather than limiting them to predefined outputs or ephemeral tools.

As technology advances, more interactive artists and designers will likely embrace open small models, benefiting from their independence and control over creative practice. This shift ensures that AI remains a medium for artistic expression and allows many types of innovative interactive interface, rather than being restricted to a corporate controlled tool prone to obsolescence. By engaging with open-source communities, artists can contribute to the expansion, refinement, and accessibility of open small AI models, playing an active role in shaping the future of reliable and artist-driven AI creativity.

For those with a hacker's spirit, especially interactive artists needing to integrate diverse technologies, small models offer a playground for experimentation, customization, and innovation, a space where AI is not merely used, but actively co-created, integrated, and redefined. This open-ended approach fosters a richer, inclusive, and diverse landscape for digital and interactive art, ensuring that AI remains a force for creative empowerment rather than creative limitation imposed by closed systems. Ultimately, open small AI models ensure that interactive artists remain the authors not only of their artworks but of the systems that shape them.

\bibliographystyle{ACM-Reference-Format}
\bibliography{references} 


\begin{thebibliography}{22}

%%% the Web address.

\ifx \showCODEN    \undefined \def \showCODEN     #1{\unskip}     \fi
\ifx \showDOI      \undefined \def \showDOI       #1{#1}\fi
\ifx \showISBNx    \undefined \def \showISBNx     #1{\unskip}     \fi
\ifx \showISBNxiii \undefined \def \showISBNxiii  #1{\unskip}     \fi
\ifx \showISSN     \undefined \def \showISSN      #1{\unskip}     \fi
\ifx \showLCCN     \undefined \def \showLCCN      #1{\unskip}     \fi
\ifx \shownote     \undefined \def \shownote      #1{#1}          \fi
\ifx \showarticletitle \undefined \def \showarticletitle #1{#1}   \fi
\ifx \showURL      \undefined \def \showURL       {\relax}        \fi
% invisible to TeX
\providecommand\bibfield[2]{#2}
\providecommand\bibinfo[2]{#2}
\providecommand\natexlab[1]{#1}
\providecommand\showeprint[2][]{arXiv:#2}

\bibitem[Abuzuraiq and Pasquier(2024a)]%
        {abuzuraiq2024seizing}
\bibfield{author}{\bibinfo{person}{Ahmad~Mohammad Abuzuraiq} {and} \bibinfo{person}{Philippe Pasquier}.} \bibinfo{year}{2024}\natexlab{a}.
\newblock \showarticletitle{Seizing the Means of Production: Exploring the Landscape of Crafting, Adapting and Navigating Generative AI Models}. In \bibinfo{booktitle}{\emph{3rd Generative AI and HCI Workshop, CHI '24}}.
\newblock


\bibitem[Abuzuraiq and Pasquier(2024b)]%
        {abuzuraiq2024towards}
\bibfield{author}{\bibinfo{person}{Ahmad~Mohammad Abuzuraiq} {and} \bibinfo{person}{Philippe Pasquier}.} \bibinfo{year}{2024}\natexlab{b}.
\newblock \showarticletitle{Towards Personalizing Generative AI with Small Data for Co-Creation in the Visual Arts}. In \bibinfo{booktitle}{\emph{Proceedings of the Workshops at the 29th International Conference on Intelligent User Interfaces (IUI '24)}} \emph{(\bibinfo{series}{CEUR Workshop Proceedings}, Vol.~\bibinfo{volume}{3660})}.
\newblock
\urldef\tempurl%
\url{https://ceur-ws.org/Vol-3660/paper11.pdf}
\showURL{%
\tempurl}


\bibitem[Broad et~al\mbox{.}(2021)]%
        {broad2021active}
\bibfield{author}{\bibinfo{person}{Terence Broad}, \bibinfo{person}{Sebastian Berns}, \bibinfo{person}{Simon Colton}, {and} \bibinfo{person}{Mick Grierson}.} \bibinfo{year}{2021}\natexlab{}.
\newblock \showarticletitle{Active Divergence with Generative Deep Learning: A Survey and Taxonomy}. In \bibinfo{booktitle}{\emph{Proceedings of the 12th International Conference on Computational Creativity (ICCC '21)}}. \bibinfo{pages}{306--315}.
\newblock
\urldef\tempurl%
\url{https://computationalcreativity.net/iccc21/wp-content/uploads/2021/09/ICCC_2021_paper_43.pdf}
\showURL{%
\tempurl}


\bibitem[Bryan-Kinns et~al\mbox{.}(2025)]%
        {bryankinns2025xaiarts}
\bibfield{author}{\bibinfo{person}{Nick Bryan-Kinns}, \bibinfo{person}{Shuoyang Zheng}, \bibinfo{person}{Francisco Castro}, \bibinfo{person}{Makayla Lewis}, \bibinfo{person}{Jia-Rey Chang}, \bibinfo{person}{Gabriel Vigliensoni}, \bibinfo{person}{Terence Broad}, \bibinfo{person}{Michael~Paul Clemens}, {and} \bibinfo{person}{Elizabeth Wilson}.} \bibinfo{year}{2025}\natexlab{}.
\newblock \showarticletitle{XAIxArts Manifesto: Explainable AI for the Arts}. In \bibinfo{booktitle}{\emph{Proceedings of the Extended Abstracts of the CHI Conference on Human Factors in Computing Systems}}. \bibinfo{pages}{1--8}.
\newblock


\bibitem[Canet~Sola and Guljajeva(2024)]%
        {canet2024visions}
\bibfield{author}{\bibinfo{person}{Mar Canet~Sola} {and} \bibinfo{person}{Varvara Guljajeva}.} \bibinfo{year}{2024}\natexlab{}.
\newblock \showarticletitle{Visions of Destruction: Exploring Human Impact on Nature by Navigating the Latent Space of a Diffusion Model via Gaze}. In \bibinfo{booktitle}{\emph{Proceedings of the Eighteenth International Conference on Tangible, Embedded, and Embodied Interaction (TEI '24)}}. \bibinfo{publisher}{Association for Computing Machinery}, \bibinfo{address}{New York, NY, USA}, \bibinfo{pages}{1--5}.
\newblock
\urldef\tempurl%
\url{https://doi.org/10.1145/3623509.3635319}
\showDOI{\tempurl}


\bibitem[Epstein et~al\mbox{.}(2023)]%
        {epstein2023art}
\bibfield{author}{\bibinfo{person}{Ziv Epstein}, \bibinfo{person}{Aaron Hertzmann}, \bibinfo{person}{the Investigators~of Human~Creativity}, \bibinfo{person}{Memo Akten}, \bibinfo{person}{Hany Farid}, \bibinfo{person}{Jessica Fjeld}, \bibinfo{person}{Morgan~R. Frank}, \bibinfo{person}{Matthew Groh}, \bibinfo{person}{Laura Herman}, \bibinfo{person}{Neil Leach}, \bibinfo{person}{Robert Mahari}, \bibinfo{person}{Alex~“Sandy” Pentland}, \bibinfo{person}{Olga Russakovsky}, \bibinfo{person}{Hope Schroeder}, {and} \bibinfo{person}{Amy Smith}.} \bibinfo{year}{2023}\natexlab{}.
\newblock \showarticletitle{Art and the science of generative AI}.
\newblock \bibinfo{journal}{\emph{Science}} \bibinfo{volume}{380}, \bibinfo{number}{6650} (\bibinfo{year}{2023}), \bibinfo{pages}{1110--1111}.
\newblock


\bibitem[Guljajeva and Canet~Sola(2022a)]%
        {guljajeva2022dream}
\bibfield{author}{\bibinfo{person}{Varvara Guljajeva} {and} \bibinfo{person}{Mar Canet~Sola}.} \bibinfo{year}{2022}\natexlab{a}.
\newblock \showarticletitle{Dream Painter: An Interactive Art Installation Bridging Audience Interaction, Robotics, and Creative AI}. In \bibinfo{booktitle}{\emph{Proceedings of the 30th ACM International Conference on Multimedia (MM '22)}}. \bibinfo{publisher}{Association for Computing Machinery}, \bibinfo{address}{New York, NY, USA}, \bibinfo{pages}{7235--7236}.
\newblock
\urldef\tempurl%
\url{https://doi.org/10.1145/3503161.354997}
\showDOI{\tempurl}


\bibitem[Guljajeva and Canet~Sola(2022b)]%
        {guljajeva2022keepsmiling}
\bibfield{author}{\bibinfo{person}{Varvara Guljajeva} {and} \bibinfo{person}{Mar Canet~Sola}.} \bibinfo{year}{2022}\natexlab{b}.
\newblock \showarticletitle{Keep Smiling}. In \bibinfo{booktitle}{\emph{Proceedings of the 10th Conference on Computation, Communication, Aesthetics \& X (xCoAx 2022)}}, \bibfield{editor}{\bibinfo{person}{Andr{\'e} Rangel}, \bibinfo{person}{Lu{\'\i}sa Ribas}, \bibinfo{person}{Mario Verdicchio}, {and} \bibinfo{person}{Miguel Carvalhais}} (Eds.). \bibinfo{publisher}{xCoAx}, \bibinfo{address}{Coimbra, Portugal}, \bibinfo{pages}{340--346}.
\newblock
\urldef\tempurl%
\url{https://doi.org/10.24840/xCoAx_2022_66}
\showDOI{\tempurl}


\bibitem[Guljajeva et~al\mbox{.}(2024)]%
        {guljajeva2024artist}
\bibfield{author}{\bibinfo{person}{Varvara Guljajeva}, \bibinfo{person}{Mar Canet~Sola}, {and} \bibinfo{person}{Isaac Clarke}.} \bibinfo{year}{2024}\natexlab{}.
\newblock \showarticletitle{Artist-guided Neural Networks—Automated Creativity or Tools for Extending Minds?}
\newblock In \bibinfo{booktitle}{\emph{Artificial Intelligence – Intelligent Art: Interdisciplinary Approaches to the Future of AI and Art}}, \bibfield{editor}{\bibinfo{person}{M.~Fleck}, \bibinfo{person}{L.~J{\"u}nger}, {and} \bibinfo{person}{L.~Jacobi}} (Eds.). \bibinfo{publisher}{transcript Verlag}, \bibinfo{address}{Bielefeld}, \bibinfo{pages}{59--77}.
\newblock
\showISBNx{978-3-8376-6922-0}
\urldef\tempurl%
\url{https://www.transcript-open.de/pdf_chapter/bis%206999/9783839469224/9783839469224-004.pdf}
\showURL{%
\tempurl}


\bibitem[Guljajeva and Sola(2022)]%
        {guljajeva2022keep}
\bibfield{author}{\bibinfo{person}{Varvara Guljajeva} {and} \bibinfo{person}{Mar~Canet Sola}.} \bibinfo{year}{2022}\natexlab{}.
\newblock \showarticletitle{Keep Smiling.}. In \bibinfo{booktitle}{\emph{SIGGRAPH Asia Art Gallery}}. \bibinfo{pages}{7--1}.
\newblock


\bibitem[HaCohen et~al\mbox{.}(2024)]%
        {hacohen2024ltx}
\bibfield{author}{\bibinfo{person}{Yoav HaCohen}, \bibinfo{person}{Nisan Chiprut}, \bibinfo{person}{Benny Brazowski}, \bibinfo{person}{Daniel Shalem}, \bibinfo{person}{Dudu Moshe}, \bibinfo{person}{Eitan Richardson}, \bibinfo{person}{Eran Levin}, \bibinfo{person}{Guy Shiran}, \bibinfo{person}{Nir Zabari}, \bibinfo{person}{Ori Gordon}, {et~al\mbox{.}}} \bibinfo{year}{2024}\natexlab{}.
\newblock \showarticletitle{Ltx-video: Realtime video latent diffusion}.
\newblock \bibinfo{journal}{\emph{arXiv preprint arXiv:2501.00103}} (\bibinfo{year}{2024}).
\newblock


\bibitem[Kapoor et~al\mbox{.}(2024)]%
        {kapoor2024societal}
\bibfield{author}{\bibinfo{person}{Sayash Kapoor}, \bibinfo{person}{Rishi Bommasani}, \bibinfo{person}{Kevin Klyman}, \bibinfo{person}{Shayne Longpre}, \bibinfo{person}{Ashwin Ramaswami}, \bibinfo{person}{Peter Cihon}, \bibinfo{person}{Aspen Hopkins}, \bibinfo{person}{Kevin Bankston}, \bibinfo{person}{Stella Biderman}, \bibinfo{person}{Miranda Bogen}, \bibinfo{person}{Rumman Chowdhury}, \bibinfo{person}{Alex Engler}, \bibinfo{person}{Peter Henderson}, \bibinfo{person}{Yacine Jernite}, \bibinfo{person}{Seth Lazar}, \bibinfo{person}{Stefano Maffulli}, \bibinfo{person}{Alondra Nelson}, \bibinfo{person}{Joelle Pineau}, \bibinfo{person}{Aviya Skowron}, \bibinfo{person}{Dawn Song}, \bibinfo{person}{Victor Storchan}, \bibinfo{person}{Daniel Zhang}, \bibinfo{person}{Daniel~E. Ho}, \bibinfo{person}{Percy Liang}, {and} \bibinfo{person}{Arvind Narayanan}.} \bibinfo{year}{2024}\natexlab{}.
\newblock \bibinfo{title}{On the Societal Impact of Open Foundation Models}.
\newblock
\newblock
\showeprint[arxiv]{2403.07918}~[cs.CY]
\urldef\tempurl%
\url{https://arxiv.org/abs/2403.07918}
\showURL{%
\tempurl}


\bibitem[Kelomees et~al\mbox{.}(2022)]%
        {kelomees2022meaning}
\bibfield{author}{\bibinfo{person}{Raivo Kelomees}, \bibinfo{person}{Varvara Guljajeva}, {and} \bibinfo{person}{Oliver Laas}.} \bibinfo{year}{2022}\natexlab{}.
\newblock \bibinfo{booktitle}{\emph{The Meaning of Creativity in the Age of AI}}.
\newblock \bibinfo{publisher}{Estonian Academy of Arts}, \bibinfo{address}{Tallinn}.
\newblock
\urldef\tempurl%
\url{https://var-mar.info/wp-content/uploads/2023/10/The-Meaning-of-Creativity-in-the-Age-of-AI.pdf}
\showURL{%
\tempurl}


\bibitem[Lozano-Hemmer(1996)]%
        {lozano1996perverting}
\bibfield{author}{\bibinfo{person}{Rafael Lozano-Hemmer}.} \bibinfo{year}{1996}\natexlab{}.
\newblock \showarticletitle{Perverting Technological Correctness}.
\newblock \bibinfo{journal}{\emph{Leonardo}} \bibinfo{volume}{29}, \bibinfo{number}{1} (\bibinfo{year}{1996}), \bibinfo{pages}{5--15}.
\newblock


\bibitem[Lozano-Hemmer(2022)]%
        {lozano2022best}
\bibfield{author}{\bibinfo{person}{Rafael Lozano-Hemmer}.} \bibinfo{year}{2022}\natexlab{}.
\newblock \bibinfo{title}{Best Practices for Conservation of Media Art from an Artist’s Perspective}.
\newblock \bibinfo{howpublished}{GitHub repository}.
\newblock
\urldef\tempurl%
\url{https://github.com/antimodular/Best-practices-for-conservation-of-media-art}
\showURL{%
\tempurl}


\bibitem[Luccioni et~al\mbox{.}(2024)]%
        {luccioni2024power}
\bibfield{author}{\bibinfo{person}{Sasha~A. Luccioni}, \bibinfo{person}{Yacine Jernite}, {and} \bibinfo{person}{Emma Strubell}.} \bibinfo{year}{2024}\natexlab{}.
\newblock \showarticletitle{Power Hungry Processing: Watts Driving the Cost of AI Deployment?}. In \bibinfo{booktitle}{\emph{Proceedings of the 2024 ACM Conference on Fairness, Accountability, and Transparency (FAccT '24)}}. \bibinfo{publisher}{Association for Computing Machinery}, \bibinfo{address}{New York, NY, USA}, \bibinfo{pages}{2578--2591}.
\newblock
\urldef\tempurl%
\url{https://doi.org/10.1145/3630106.3658542}
\showDOI{\tempurl}


\bibitem[Manovich(2022)]%
        {manovich2022ai}
\bibfield{author}{\bibinfo{person}{Lev Manovich}.} \bibinfo{year}{2022}\natexlab{}.
\newblock \showarticletitle{AI and Myths of Creativity}.
\newblock \bibinfo{journal}{\emph{Architectural Design}} \bibinfo{volume}{92}, \bibinfo{number}{3} (\bibinfo{year}{2022}), \bibinfo{pages}{60--65}.
\newblock
\urldef\tempurl%
\url{https://doi.org/10.1002/ad.2814}
\showDOI{\tempurl}


\bibitem[McDonald(2024)]%
        {mcdonald2024realtime}
\bibfield{author}{\bibinfo{person}{Kyle McDonald}.} \bibinfo{year}{2024}\natexlab{}.
\newblock \bibinfo{title}{Realtime Diffusion in the Cloud}.
\newblock
\newblock
\urldef\tempurl%
\url{https://kcimc.medium.com/realtime-diffusion-in-the-cloud-f497e66d4f92}
\showURL{%
\tempurl}
\newblock
\shownote{Medium article, accessed November 12, 2024}.


\bibitem[Meyer(2024)]%
        {Meyer2024}
\bibfield{author}{\bibinfo{person}{Roland Meyer}.} \bibinfo{year}{2024}\natexlab{}.
\newblock \showarticletitle{It's a flat world. The Synthetic Realities of Sora}.
\newblock  \bibinfo{volume}{Special Issue 1 (2024)}, \bibinfo{number}{Special Issue 1,3} (\bibinfo{year}{2024}), \bibinfo{pages}{3}.
\newblock
\urldef\tempurl%
\url{https://doi.org/10.57684/COS-1267}
\showDOI{\tempurl}


\bibitem[Shin et~al\mbox{.}(2025)]%
        {shin2025motionstreamrealtimevideogeneration}
\bibfield{author}{\bibinfo{person}{Joonghyuk Shin}, \bibinfo{person}{Zhengqi Li}, \bibinfo{person}{Richard Zhang}, \bibinfo{person}{Jun-Yan Zhu}, \bibinfo{person}{Jaesik Park}, \bibinfo{person}{Eli Schechtman}, {and} \bibinfo{person}{Xun Huang}.} \bibinfo{year}{2025}\natexlab{}.
\newblock \bibinfo{title}{MotionStream: Real-Time Video Generation with Interactive Motion Controls}.
\newblock
\newblock
\showeprint[arxiv]{2511.01266}~[cs.CV]
\urldef\tempurl%
\url{https://arxiv.org/abs/2511.01266}
\showURL{%
\tempurl}


\bibitem[Tecks et~al\mbox{.}(2024)]%
        {tecks2024explainability}
\bibfield{author}{\bibinfo{person}{Austin Tecks}, \bibinfo{person}{Thomas Peschlow}, {and} \bibinfo{person}{Gabriel Vigliensoni}.} \bibinfo{year}{2024}\natexlab{}.
\newblock \showarticletitle{Explainability Paths for Sustained Artistic Practice with AI}. In \bibinfo{booktitle}{\emph{Proceedings of the Explainable AI for the Arts Workshop (XAIxArts 2024)}}.
\newblock
\urldef\tempurl%
\url{https://doi.org/10.48550/arXiv.2407.15216}
\showDOI{\tempurl}


\bibitem[Vigliensoni et~al\mbox{.}(2022)]%
        {vigliensoni2022small}
\bibfield{author}{\bibinfo{person}{Gabriel Vigliensoni}, \bibinfo{person}{Phoenix Perry}, {and} \bibinfo{person}{Rebecca Fiebrink}.} \bibinfo{year}{2022}\natexlab{}.
\newblock \showarticletitle{A Small-Data Mindset for Generative AI Creative Work}. In \bibinfo{booktitle}{\emph{Workshop on Generative AI and HCI at CHI '22}}.
\newblock
\urldef\tempurl%
\url{https://ualresearchonline.arts.ac.uk/id/eprint/18343/}
\showURL{%
\tempurl}


\end{thebibliography}

\end{document}